\documentclass[applsci,article,accept,moreauthors,pdftex]{Definitions/mdpi}
\pdfoutput=1

\firstpage{1} 
\pubvolume{1}
\issuenum{1}
\articlenumber{0}
\pubyear{2021}
\copyrightyear{2021}
\datereceived{31 October 2020} 
\dateaccepted{4 January 2021} 
\datepublished{} 
\hreflink{https://doi.org/} 

\usepackage{graphicx} 
\usepackage[caption=false,font=footnotesize]{subfig}

\usepackage[ruled,vlined]{algorithm2e}
\usepackage{setspace}
\usepackage{hyperref}
\usepackage{amsthm}
\usepackage{amsmath}
\usepackage{amssymb}
\usepackage{multirow}
\usepackage{rotating}
\usepackage{xspace}
\usepackage{courier}
\usepackage{listings}

\usepackage{easybmat}
\usepackage{color}
\usepackage{enumitem}
\usepackage{cancel}

\newcommand{\mathds}[1]{\mathbb{#1}}
\newcommand{\beq}{\begin{equation}}
\newcommand{\eeq}{\end{equation}}
\newcommand{\R}{\mathds{R}}

\newcommand{\skl}[1]{{#1}}

\DeclareMathOperator*{\argmax}{arg\,max}

\usepackage{bbm}

\usepackage{diagbox}
\usepackage{slashbox}

\usepackage{multirow}

\usepackage{subfiles}
\usepackage{xr}
\usepackage{import}

\Title{Data Quality Measures and Efficient Evaluation Algorithms for Large-Scale High-Dimensional Data}

\TitleCitation{Data Quality Measures and Efficient Evaluation Algorithms for Large-Scale High-Dimensional Data}

\Author{Hyeongmin Cho $^{1}$ and Sangkyun Lee $^{1,}$*}

\AuthorNames{Hyeongmin Cho and Sangkyun Lee}
\AuthorCitation{Cho, H.; Lee, S.}
\address{%
$^{1}$ \quad School of Cybersecurity, Korea University, Seoul 02841, Korea\\} 

\corres{Correspondence: sangkyun@korea.ac.kr}

\abstract{
  Machine learning has been proven to be effective in various application areas, such as object and speech recognition on mobile systems. Since a critical key to machine learning success is the availability of large training data, many datasets are being disclosed and published online. From a data consumer or manager point of view, measuring data quality is an important first step in the learning process. We need to determine which datasets to use, update, and maintain. However, not many practical ways to measure data quality are available today, especially when it comes to large-scale high-dimensional data, such as images and videos. This paper proposes two data quality measures that can compute class separability and in-class variability, the two important aspects of data quality, for a given dataset. Classical data quality measures tend to focus only on class separability; however, we suggest that in-class variability is another important data quality factor. We provide efficient algorithms to compute our quality measures based on random projections and bootstrapping with statistical benefits on large-scale high-dimensional data. In experiments, we show that our measures are compatible with classical measures on small-scale data and can be computed much more efficiently on large-scale high-dimensional datasets.}


\begin{document}


\section{Introduction}
We are witnessing the success of machine learning in various research and application areas, such as vision inspection, energy consumption estimation, and autonomous driving, just to name a few. One major contributor to the success is the fact that the datasets are continuously accumulated and openly published in several domains. Low data quality is very likely to cause inferior prediction performance of machine learning models, and therefore measuring data quality is an indispensable step in a machine learning process. Especially in real-time and mission-critical cyber-physical-system, defining appropriate data quality measures is critical since the low generalization performance of a deployed model can result in system malfunction and possibly catastrophic damages to the physical world. Despite the importance, there exist only a few works for measuring data quality where most of them are hard to evaluate on large-scale high-dimensional data due to computation complexity.

A popular early work on data quality measures includes \citet{Ho_2002}, proposing 12 types of quality measures which are simple but powerful enough to address different aspects of data quality. These measures have limitations, however, in that it is difficult to compute them for large-scale high-dimensional and multi-class datasets. \citet{Baumgartner_2006} proposed a quality measure designed for high-dimensional biomedical datasets; however, it does not work efficiently on large-scale data. Recently, Branchaud-Charron et al. \cite{Branchaud_2019} proposed a quality measure for high-dimensional data using spectral clustering. Although this measure is adequate for large-scale high-dimensional data, it requires an embedding network which involves a large amount of computation time for~training. 

In this paper, we propose three new quality measures called $M_{sep}$, $M_{var}$, and $M_{var_i}$, and their computation methods that overcome the above limitations. Our approach is inspired by Fisher's linear discriminant analysis (LDA)~\cite{Li_2014}, which is mathematically well-defined for finding a feature subspace that maximizes class separability. Our computation method makes use of the techniques from statistics, random projection~\cite{Bingham_2001} and bootstrapping~\cite{Efron_1979}, to compute the measure for large-scale high-dimensional data efficiently.

The contributions of our paper are summarized as follows:
\begin{itemize}
    \item \skl{We propose three new data quality measures that can be compuated directly from a given dataset and can be compared across datasets with different numbers of classes, examples, and features. Although our approach takes ideas from LDA (linear discriminant analysis) and PCA (principal component analysis), the techniques by themselves do not produce single numbers that are comparable across different datasets. }

    \item  We provide efficient algorithms to approximate the suggested data quality measures, making them available for large-scale high-dimensional data.


    \item The proposed class separability measure $M_{sep}$ is strongly correlated with the actual classification performance of linear and non-linear classifiers in our experiments.
    
    \item \skl{The proposed in-class variability measures $M_{var}$ and $M_{var_i}$ quantify the diversity of data within each class and can be used to analyze redundancy or outlier issues}.


\end{itemize}


\section{Related Work}
In general, the quality of data can be measured by Kolmogorov complexity which is also known as the descriptive complexity for algorithmic entropy~\cite{MingLi_2008}. However, the Kolmogorov complexity is not computationally feasible; instead, an approximation is used in practice. To our knowledge, there are three main approaches for approximating the Kolmogorov complexity: descriptor-based, classifier-based, and graph-based approaches. We describe these three main categories below.

\subsection{Descriptor-Based Approaches} \label{subsec:descriptor}
\citet{Ho_2002} proposed simple but powerful quality measures based on descriptors. They proposed 12 quality measures, namely $F1$, $F2$, $F3$, $L1$, $L2$, $L3$, $N1$, $N2$, $N3$, $N4$, $T1$, and $T2$. The $F$ measures represent the amount of feature overlap. In particular, $F1$ measures the maximum Fisher's ratio which represents the maximum discriminant power of between features. $F2$ represents the volume of overlap region in the two-class conditional distributions.~$F3$ captures the ratio of overlapping features using the maximum and minimum value. The $L$ measures are for the linear separability of classes. $L1$ is a minimum error of linear programming (LP), $L2$ is an error rate of a linear classifier by LP, and $L3$ is an error rate of linear classifier after feature interpolation. The $N$ measures represent mixture identifiability, the distinguishability of the data points belonging to two different classes. $N1$ represents the ratio of nodes connected to the different classes using the minimum spanning tree of all data points. $N2$ is the ratio of the average intra-class distance and average inter-class distance. $N3$ is the leave-one-out error rate of the nearest neighbor (1NN). $N4$ is an error rate of 1NN after feature interpolation. The $T$ measure represents the topological characteristic of a dataset. $T1$ represents the number of hyperspheres adjacent to other class features. $T2$ is the average number of data points per dimension. These quality measures can capture various aspects of data quality; however, they are fixed for binary classification and not applicable for multi-class problems. Furthermore, quality measures, such as $N1$, $N2$, and $N3$, require a large amount of computation time on large-scale high-dimensional~data.

\citet{Baumgartner_2006} proposed a quality measure for high-dimensional but small biomedical datasets. They used singular value decomposition (SVD) with time complexity $\mathcal O(min(m^2n, mn^2))$, where $m$ is the number of data points, and $n$ is the number of features. Thus, it is computationally demanding to calculate their measures for the datasets with large $m$ and $n$, such as recent image datasets.

There are other descriptor-based approaches, for example for meta learning~\cite{Leyva_2015,  Sotoca_2006}, for classifier recommendation~\cite{Garcia_2018}, and for synthetic data generation~\cite{Melo_2018}.~However, only a small number of data points in a low dimensional space have been considered in these~works.

\subsection{Graph-Based Approaches} \label{subsec:graph}
Branchaud-Charron et al. \cite{Branchaud_2019} proposed a graph-based quality measure using spectral clustering. First, they compute a probabilistic divergence-based $K \times K$ class similarity matrix $S$, where $K$ is the number of classes. Then, an adjacency matrix $W$ is computed from the $S$ matrix. The quality measure is defined as a cumulative sum of the eigenvalues gap which is called as cumulative spectral gradient ($CSG$), which represents the minimum cutting cost of the $S$. The authors also used a convolutional neural network-based auto-encoder and t-SNE~
\cite{Masci_2011, Maaten_2008} to find an embedding that can represent data points (images in their case) well. Although the method is designed for high-dimensional data, it requires to train a good embedding network to reach quality performance. 

\citet{Duin_2006} proposed a quality measure based on a dissimilarity matrix of data points. Since calculating the dissimilarity matrix is a time-consuming process, the method is not adequate for large-scale high-dimensional data.

\subsection{Classifier-Based Approaches}
Li et al. \cite{Li_2018} proposed a classifier-based quality measure called an intrinsic dimension, which is the minimum number of solutions for certain problems. For example, in a neural network, the intrinsic dimension is the minimum number of parameters to reach the desired prediction performance. 

The method has a benefit that it can be applied to many different types of data as long as one has trainable classifiers for the data; however, it often incurs high computation cost since it needs to change the number of classifier parameters iteratively during data quality~evaluation.

Overall, the existing data quality measures are mostly designed for binary classification in low dimension spaces with a small number of data points. Due to their computation complexity, they tend to consume large amount of time when applied to large-scale high-dimensional data. In addition, the existing measures tend to focus only on the inter-class aspects of data quality. In this paper, we propose two new data quality measures suitable for large-scale high-dimensional data resolving the above mentioned issues.


\section{Methods}
In this section, we formally describe our data quality measures. We focus on multi-class classification tasks where each data point is associated with a class label out of $c$~categories ($c\ge 2$). Our measures are created by adapting ideas from Fisher's LDA~\cite{Li_2014}. Fisher's LDA is a dimensionality reduction technique, finding a projection matrix that maximizes the between-class variance and minimizes the within-class variance at the same time. Motivated by the idea, we propose two types of data quality measures, class separability $M_{sep}$ and in-class variability $M_{var}$ and $M_{var_i}$. For efficient handling of large-scale high-dimensional data, we also propose techniques to reduce both computation and memory requirements taking advantage of statistical methods, bootstrapping~\cite{Efron_1979} and random projection~\cite{Bingham_2001}.

\subsection{Fisher's LDA}
The objective of Fisher's LDA~\cite{Li_2014} is to find the feature subspace which maximizes the linear separability of a given dataset. Fisher's LDA achieves the objective by minimizing the within-class variance and maximizing the between-class variance simultaneously.

To describe the Fisher's LDA formally, let us consider an input matrix $X\in\mathbb{R}^{m\times n}$ where $m$ is the number of data points, $n$ is the input dimension, and $x_{i,j}\in\mathbb{R}^n$ is a $j$-th data point in the $i$-th class. The within-class scatter matrix $S_w\in\mathbb{R}^{n\times n}$ is defined as follows:
\begin{equation*}
S_w = \sum^c_{i=1}\sum^{m_i}_{j=1}(x_{i,j}-\overline{x}_i)(x_{i,j}-\overline{x}_i)^T. \enspace
\end{equation*}

Here, $c$ is the number of classes, $m_i$ is the number of data points in the $i$-th class, and $\overline{x}_i\in\mathbb{R}^{n}$ is the mean of data points in the $i$-th class. This formulation can be interpreted as the sum of class-wise scatter matrices. A small determinant of $S_w$ indicates that data points in the same class exist densely in a narrow area, which may lead to high class separability.

Next, the between-class scatter matrix $S_b\in\mathbb{R}^{n\times n}$ is defined as follows:
\begin{equation*}
S_b = \sum_{i=1}^c m_i \left(\overline{x}_i - \overline{x}\right) \left(\overline{x}_i - \overline{x}\right)^T, \enspace
\end{equation*}
where $\overline{x}$ is the mean of entire data points in the given dataset. A large determinant of $S_b$ indicates that the mean vector $\overline{x}_i$ of each class is far from the $\overline{x}$, another condition hinting for high class separability.

Using these two matrices, we can describe the objective of Fisher's LDA as follows:
\begin{equation}\label{eq:LDA}
    \Phi_{lda} \in \argmax_{\Phi}\frac{|\Phi^T S_b \Phi|}{|\Phi^T S_w \Phi|}. \enspace
\end{equation}

Here, $\Phi_{lda}\in \mathbb{R}^{n\times d}$ is the projection matrix where $d$ is the dimension of feature subspace (in general, we choose $d \ll n$). The column vectors of projection matrix $\Phi_{lda}$ are the axes of feature subspace, which maximize the class separability. 
The term in the objective function is also known as the Fisher's criterion.
By projecting $X$ onto these axes, we obtain a $d$-dimensional projection of the original data $X^{\prime}\in\mathbb{R}^{m\times d}$:
\begin{equation*}
    X^{\prime} = X\Phi_{lda}. \enspace
\end{equation*}

In general, if $S_w$ is an invertible matrix, we can calculate the projection matrix which maximizes the objective of the Fisher's LDA by eigenvalue decomposition.

\subsection{Proposed Data Quality Measures}\label{sec:proposed_measures}
Motivated by the ideas in Fisher's LDA, we propose two types of new data quality measures: $M_{sep}$ (class separability), $M_{var}$ and $M_{var_i}$ (in-class variability).

\subsubsection{Class Separability}
Our first data quality measure tries to capture the class separability of a dataset by combining the within-class variance and between-class variance, similarly to Fisher's LDA~\eqref{eq:LDA} but more efficiently for large-scale and high-dimensional data and comparable with other datasets. 

We start from creating the normalized versions of the matrices $S_w$ and $S_b$ in Fisher's LDA~\eqref{eq:LDA} so that they will not be affected by the different numbers of examples in classes ($m_i$ is the number of examples in the $i$-th class) across different datasets. The normalized versions are denoted by $\hat{S}_w$ and $\hat{S}_b$:
\begin{equation}\label{eq:normalized_components}
    \hat{S}_{w} := \sum^{c}_{i=1}\frac{1}{m_i}\sum^{m_i}_{j=1}(x_{i,j}-\overline{x}_i)(x_{i,j}-\overline{x}_i)^T , \qquad 
    \hat{S}_{b} := \sum^{c}_{i=1}\frac{m_i}{\sum^{c}_{j=1}m_j}(\overline{x}_{i}-\overline{x})(\overline{x}_{i}-\overline{x})^T. \enspace
\end{equation}

Considering the determinants of these $n \times n$ matrices as in the original Fisher's LDA will be too costly for a high-dimensional data where $n$ is large, since the time complexity to compute the determinants will be proportional nearly to $n^3$. Instead, we consider the direction of maximum linear separation $v \in \R^n$ that maximizes the ratio of between-class variance to the within-class variance being projected onto the vector. Using the vector, we define our first data quality measure $M_{sep}$ for class separability as follows:

\begin{equation}\label{eq:I_L}
    M_{sep} := \max_{v \in \R^n: \|v\|=1}\frac{|v^T \hat{S}_b v|}{|v^T \hat{S}_w v|}. \enspace
 \end{equation}
 
This formulation is almost the same as \eqref{eq:LDA} in Fisher's LDA except that \eqref{eq:LDA} finds the projection matrix $\phi_{lda}$ which maximizes the Fisher's criterion, while, in \eqref{eq:I_L}, we will focus on finding the maximum value of Fisher's criterion itself. Unlike Fisher's criterion, our measure $M_{sep}$ is comparable across datasets with different numbers of classes and examples due to normalization, to check the relative difficulty of linear classification.

Solving \eqref{eq:I_L} directly will be preventive for a large $n$ as in the original LDA. If $\hat{S}_w$ is invertible, we can calculate the vector which maximizes $M_{sep}$ as follows using simple linear algebra. To find the vector $v$ which maximizes the equation in \eqref{eq:I_L}, first differentiate it with respect to $v$ to get:
\begin{equation*}
    (v^{T} \hat{S}_b v)\hat{S}_w v = (v^{T} \hat{S}_w v)\hat{S}_b v. \enspace
\end{equation*}

This leads us to the following generalized eigenvalue problem in the form of:
\begin{equation}\label{eq:eigen_value}
    \hat{S}_w^{-1} \hat{S}_b v=\lambda v, \enspace
\end{equation}
where $\lambda=\frac{v^{T} \hat{S}_b v}{v^{T} \hat{S}_w v}$ can be thought as an eigenvalue of the matrix $\hat{S}_w^{-1} \hat{S}_b$. The maximizer $v$ is the eigenvector corresponding to the largest eigenvalue of $\hat{S}_w^{-1} \hat{S}_b$ which can be found rather efficiently by the Lanczos algorithm~\cite{Golub_1996}. However, the overall time complexity for computation can be up to $\mathcal{O}(n^3)$, which makes it difficult to calculate the optimal vector for high-dimensional data, such as images. In Section~\ref{sec:compute}, we provide an efficient algorithm to compute $M_{sep}$ using random projection.


\subsubsection{In-Class Variability}
\begin{figure}[H]
    \includegraphics[width=0.45\linewidth]{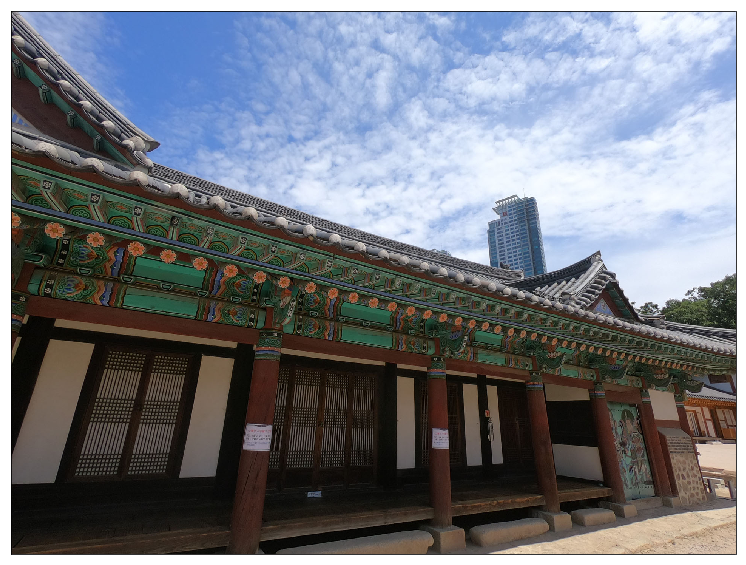}
    \includegraphics[width=0.45\linewidth]{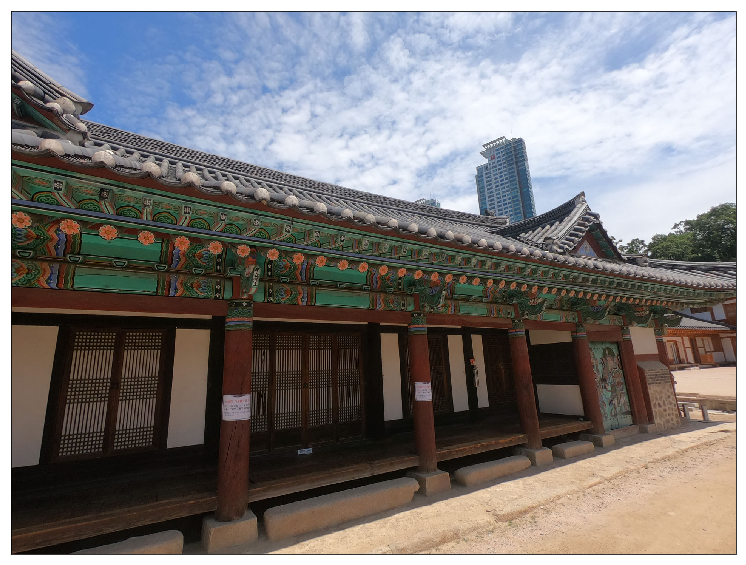}
    \caption{An example of low in-class variability that similar images in the same class. The images are Bongeunsa temple in Seoul, Korea. (Source: Korean Type Object Image AI Training Dataset at \url{http://www.aihub.or.kr/aidata/132}, National Information Society Agency.)}
    \label{fig:temple}
\end{figure}

Our second data quality measure gauges the in-class variability. Figure~\ref{fig:temple} shows one of the motivating examples to consider in-class variability for data quality. In the figure, we have two photos of the Bongeunsa temple in Seoul, Korea, taken by the same photographer. The photographer had been asked to take photos of Korean objects from several different angles, and it turned out that quite a few of the photos were taken in only marginal angle differences. Since the data creation was a government-funded project providing data for developing object recognition systems in academia and industry, low data variability was definitely an issue.

Here, we define two types of in-class variability measure, \skl{the overall in-class variability of a given dataset $M_{var}$ and the in-class variability of the $i$-th class, $M_{var_i}$. First, the overall in-class variability $M_{var}$ 
tries to capture the minimum variance of data points being projected onto any direction, 
based on the matrix $\hat{S}_{w}$ defined in \eqref{eq:normalized_components}: }
\begin{equation*}
    M_{var} := \min_{v \in \R^n: \|v\|=1} \frac{1}{c \cdot n}  v^T \hat{S}_{w} v, \enspace
\end{equation*}
where $c$ is the number of class and $n$ is the dimension of data. Unlike class separability, we added additional normalization factors $c$ and $n$, since the value $\hat{S}_{w}$ is affected by the number of class and the data dimension. 

\skl{Second, the class-wise in-class variability $M_{var_i}$
is based on the sample covariance matrix of each class:}

\begin{equation*}
    \hat{S}_{w_i} := \frac{1}{m_i}\sum^{m_i}_{j=1}(x_{i,j}-\overline{x}_i)(x_{i,j}-\overline{x}_i)^T, \enspace
\end{equation*}
\skl{where $m_i$ is the number of data points in the $i$-th class.} 
The class-wise in-class variability measure $M_{var_i}$ is defined as follows:

\begin{equation*}
M_{var_i} := \min_{v \in \R^n: \|v\|=1} \frac{1}{c \cdot n} v^T \hat{S}_{w_i} v \enspace.
\end{equation*}

\skl{The normalization factors $c$ and $n$ are required for the same reason as in $M_{var}$.} The measure $M_{var_i}$ represents the smallest variance of the data points in the same class after being projected onto any direction.

As a matter of fact, $M_{var}$ and $M_{var_i}$ are the same as the smallest eigenvalue of $\frac{1}{c \cdot n} \hat{S}_{w}$ and $\frac{1}{c \cdot n} \hat{S}_{w_i}$ which can be computed for instance using the Lanczos algorithm~\cite{Golub_1996} on the inverse of them with $\mathcal O(n^3)$ computation, which will be preventive for large data dimensions $n$. We discuss a more efficient way to estimate the value in the next section, \skl{which can be computed alongside with our first data quality measure without significant extra cost.}

Using the $M_{var}$ and $M_{var_i}$, we can analyze the variety or redundancy in a given dataset. For instance, a very small $M_{var}$ and $M_{var_i}$ would indicate that we may have a small diversity issue where data points in the invested class are mostly alike. On the other hand, the overly large $M_{var}$ and $M_{var_i}$ may indicate a noise issue in the class, possibly including incorrect labeling. The difference between $M_{var}$ and $M_{var_i}$ is that the $M_{var}$ aggregates the information of diversity for each class, and the $M_{var_i}$ represents the information of diversity for a specific class. Since the $M_{var}$ aggregates the information of variability of data points in each class, we can use this for comparing the in-class variability between datasets. On the other hand, we can use $M_{var_i}$ for the datasets analysis, i.e., data points of a specific class with less $M_{var_i}$ than other classes may cause low generalization performance. We will discuss more details in Section~\ref{sec:exp}.

\subsection{Methods for Efficient Computation}\label{sec:compute}
One of the key properties required for data quality measures is that they should be computable in a reasonable amount of time and computation resources since the amount and the dimension of data are keep increasing as new advanced sensing technologies become available. In this section, we describe how we avoid a large amount of time and memory complexity to compute our suggested data quality measures.

\subsubsection{Random Projection} 
Random projection~\cite{Bingham_2001} is a dimension reduction technique that can transform an $n$-dimensional vector into a $k$-dimensional vector ($k \ll n$), while preserving the critical information of the original vector. The idea behind of random projection is the Johnson-Lindenstrauss lemma~\cite{Dasgupta_1999}. That is, for any vectors $\{x, x'\} \in X$ from a set of $m$ vectors in $X \subset \R^n$ and for $\epsilon \in (0,1)$, there exists a linear mapping $f: \R^n \to \R^k$ such that the pairwise distances of vectors are almost preserved after projection in the sense that:
$$
 (1-\epsilon) \| x - x' \|^2_2 \le \|f(x) - f(x') \|^2_2 \le (1+\epsilon) \|x-x'\|^2_2,
$$
where $k > 8\ln(m)/\epsilon^2$. It is known that when the original dimension $n$ is large, a random projection matrix $P \in \R^{k \times n}$ can serve as the feature mapping $f$ in the lemma, since random vectors in $\R^n$ tend to be orthogonal to each other as $n$ increases~\cite{Kaski_1998}. 



Motivated by the above phenomenon, we use random projection to find a vector that satisfies \eqref{eq:I_L} instead of calculating the eigenvalue decomposition to solve \eqref{eq:eigen_value}.
The idea is that if the number of random vectors is sufficiently large, the maximum value of the Fisher's criterion calculated by random projection can approximate the behavior of a true solution.

Furthermore, random projection makes it unnecessary to explicitly store $\hat S_w$ and $\hat S_b$ since we can simply compute the denominator and numerator of \eqref{eq:I_L} as follows:

$$
w^T \hat{S}_{w} w = \sum^{c}_{i=1}\frac{1}{m_i}\sum^{m_i}_{j=1}w^T(x_{i,j}-\overline{x}_i)(x_{i,j}-\overline{x}_i)^T w, \enspace
$$

$$
w^T \hat S_b w = \sum_{i=1}^c\frac{m_i}{\sum_{j=1}^c m_j} w^T\left(\overline{x}_i - \overline{x}\right) \left(\overline{x}_i - \overline{x}\right)^T w, \enspace
$$
where $w$ is a random unit vector drawn from $\mathcal{N}(0,1)$. This technique is critical for dealing with high-dimensional data, such as images, in a memory-efficient way. In our experiments, ten random projection vectors were sufficient in most cases to accurately estimate our quality measures.

\subsubsection{Bootstrapping}
Bootstrapping~\cite{Efron_1979} is a sampling-based technique that estimates the statistic of the population with little data using sampling with replacement. For instance, bootstrapping can be used to estimate the mean and the variance of a statistic from an unknown population. Let $s_i$ is a statistic of interest that is calculated from a randomly drawn sample of an unknown population. The mean and variance of the statistic can be estimated as follows:
\begin{equation*}
    \hat \mu = \frac{1}{B} \sum_{i=1}^{B} s_i, \enspace \quad 
    \hat \sigma^2 = \frac{1}{B} \sum_{i=1}^{B} s_i^2 - \left( \frac{1}{B}\sum_{i=1}^{B} s_i \right)^2, \enspace
\end{equation*}
where $B$ is the number of bootstrap samples, $\hat\mu$ is a mean estimate of the statistic, and $\hat\sigma^2$ is the variability of the estimate. By using a small $B$, we can reduce the number of data points to be considered at once.
We found that $B=100$ and making each bootstrap sample to be $25\%$ of a given dataset in size worked well overall our experiments.
We summarized the above procedure in Algorithm~\ref{algorithm} ({The implementation is available here: \url{https://github.com/Hyeongmin-Cho/Efficient-Data-Quality-Measures-for-High-Dimensional-Classification-Data}.})


\newpage
\begin{algorithm}[h]
\caption{Algorithm of class separability and in-class variability.}
 \label{algorithm}
\SetAlgoLined
\KwResult{$M_{sep}$, $M_{var}$ and $M_{var_i}$ score}
 $\textbf{Dataset}=\{(x_1, y_1), \dots,(x_m, y_m)\}$ \\
 
 $\textbf{Args}$= the number of samples $B$, a sample ratio of each bootstrap sample against a given dataset $R$, the number of random vector used in each sample $nv$, an array storing the values of overall in-class variability $A_{var}$, an array storing the values of class-wise in-class variability $A_{var_i}$ and an array storing the values of class separability $A_{sep}$.\\

'$\xleftarrow{}$' symbol stands for variable assignment\\
\smallskip
$i \xleftarrow{} 1$ \\

$A_{var} \xleftarrow{} \{\}$ \\

$A_{var_i} \xleftarrow{} \{\}$ \\

$A_{sep} \xleftarrow{} \{\}$ \\

\While{i $\leq$ B}{
 $j \xleftarrow{} 1$ \\
 
 $i \xleftarrow{} i+1$ \\
 
  Calculate the number of class $c$ and data dimension $n$ from the dataset \\
 
  Sampling with replacement using stratified sampling as much as R ratio from the dataset \\

  Standardize the sampled dataset \\
  
  \While{j $\leq$ nv}{
    $j \xleftarrow{} j+1$ \\
    $w \xleftarrow{}$ a unit vector drawn from $\mathcal{N}(0,1)$ \\
    
    Compute $w^T \hat{S}_{w_{j,c}} w$ \\
    
    $w^T \hat{S}_{w_j} w \xleftarrow{} Sum(\{w^T \hat{S}_{w_{j,1}} w,\dots,w^T \hat{S}_{w_{j,c}} w\})$ \\
    $M_{var-j} \xleftarrow{} w^T \hat{S}_{w_j} w$
  
    Compute $w^T \hat{S}_{b_j} w$ \\
  
    $M_{sep-j} \xleftarrow{} (w^T \hat{S}_{b_j} w) / (w^T \hat{S}_{w_j} w)$
    
  }
    \textls[-30]{$A_{var_i}$.insert($\{\frac{1}{c \cdot n} \min(w^T \hat{S}_{w_{1,1}} w, \dots, w^T \hat{S}_{w_{nv,1}} w), \dots, \frac{1}{c \cdot n} \min(w^T \hat{S}_{w_{1,c}} w, \dots, w^T \hat{S}_{w_{nv,c}} w) \}$)} \\
    
    $A_{var}$.insert($\frac{1}{c \cdot n}\min(M_{var-1},\dots, M_{var-nv})$) \\
    
    $A_{sep}$.insert($\max(M_{sep-1},\dots, M_{sep-nv})$)
 }
 
$M_{sep} \xleftarrow{} Mean(A_{sep})$ \\
 
 $M_{var} \xleftarrow{} Mean(A_{var})$ \\
 
$M_{var_i} \xleftarrow{} ClassWiseMean(A_{var_i})$ \\

\textbf{Return} $M_{sep}, M_{var}, M_{var_i}$

\end{algorithm}




\vspace{-12pt}

\section{Experiment Results} \label{sec:exp}

In this section, we show that our method can evaluate the data quality of the large-scale high-dimensional dataset efficiently. 

To verify the representative performance of $M_{sep}$ for class separability, we calculated the correlation between the accuracy of chosen classifiers and $M_{sep}$. Classifiers used in our experiments are as follows: a perceptron, a multi-layer perceptron with one hidden layer and LeakyReLU (denoted by MLP-1), and a multi-layer perceptron with two hidden layers and LeakyReLU (denoted by MLP-2). To simplify the experiments, we trained the models with the following settings: 30 epochs, a batch size of 100, a learning rate of 0.002, the Adam optimizer, and the cross-entropy loss function. Additionally, we fixed the hyperparameters of Algorithm~\ref{algorithm} as $B=100$, $R=0.25$, and $nv=10$ since there was no big difference in performance when larger hyperparameter values were used. 

For comparison with other quality measures, we chose $F1$, $N1$, and $N3$ from~\citet{Ho_2002} and $CSG$ from~Branchaud-Charron et al. \cite{Branchaud_2019}. Here, $N1$, $N3$, and $CSG$ are known to be highly correlated with test accuracy of classifiers~Branchaud-Charron et al. \cite{Branchaud_2019}. $F1$ is similar to our $M_{sep}$ in its basic idea. Other quality measures suggested in \citet{Ho_2002} showed very similar characteristics to $F1$, $N1$, $N3$, and $CSG$ and are therefore not included in the results.



\subsection{Datasets}


To evaluate the representative performance of $M_{sep}$ for class separability, we used various image datasets that are high-dimensional and popular in mobile applications. We chose ten benchmark image datasets for our experiments: MNIST
, notMNIST, CIFAR10
, Linnaeus, STL10
, SVHN
, ImageNet-1, ImageNet-2, ImageNet-3, and ImageNet-4. MNIST~\cite{lecun_2010} consists of ten handwritten digits from 0 to 9. The dataset contains 60,000 training and 10,000~test data points. We sampled 10,000 data from the training data for a model training and measuring the quality, and we sampled 2500 data from the test data for assessing the model accuracy. The notMNIST~\cite{Bulatov_2011} dataset is quite similar to MNIST, containing English letters from A to J in various fonts. It has 13,106 training and 5618 test samples. We sampled the data in the same way as MNIST. Linnaeus~\cite{Chaladze_2017} consists of five classes: berry, bird, dog, flower, and others. Although the dataset is available in various image sizes, we chose $32 \times 32$ to reduce the computation time of $N1$, $N3$, and $CSG$. CIFAR10~\cite{Krizhevsky_2009} is for object recognition with ten general object classes. It consists of 50,000 training data and 10,000 test data points. We sampled the CIFAR10 dataset in the same way as MNIST. STL10~\cite{Coates_2011} is also for object recognition with ten classes, and it has $92 \times 92$ images: we resized the images into $32 \times 32$ to reduce the computation time for the $N1$, $N3$, and $CSG$. The dataset consists of 5000 training and 8000 test data points. We combined these two sets into a single dataset, and then sampled 10,000 data points from the combined set for model training and measuring quality. We also sampled 2500 data points from the combined set for assessing prediction model accuracy if necessary. SVHN~\cite{:Netzer} consists of street view house number images. The dataset contains 73,200 data points. We sampled 10,000 training data for a model training and measuring the quality, and we sampled 2500 data for assessing the model accuracy. ImageNet-1, ImageNet-2, ImageNet-3 and ImageNet-4 are subsets of Tiny ImageNet dataset~\cite{:Deng}. The Tiny ImageNet dataset contains 200 classes, and each class has 500 images. They are consist of randomly selected ten classes of the Tiny ImageNet dataset (total 5000 data points). We used 4500 data points for model training and measuring the quality and 500 data points for assessing the model accuracy, respectively.

We summarized the details of datasets in Table~\ref{datasets}. The accuracy values in the Table~\ref{datasets} are calculated from the MLP-2 model since it showed good overall performance compared to the perceptron and the MLP-1 models.

\begin{specialtable}[H]
\caption{Details of the datasets used in our experiments. The accuracy is from MLP-2
, and $M$ represents the total number of data used for training and evaluation.}
\setlength{\tabcolsep}{1mm}
\begin{tabular}{ccccc}
\toprule
\textbf{Datasets} & \textbf{Accuracy}   & \boldmath{$M$} & \textbf{No. Classes}     & \textbf{Description}        \\ \midrule
MNIST    & 92.64\% & 12.5 k & 10 & Hand written digit                \\
notMNIST & 89.24\% & 12.5 k & 10 & Fonts and glyphs similar to MNIST        \\
Linnaeus & 45.50\% & 4.8 k  & 5 & Botany and animal class images        \\ 
CIFAR10  & 42.84\% & 12.5 k & 10 & Object recognition images        \\
STL10    & 40.88\% & 12.5 k & 10 & Object recognition images        \\
SVHN    & 45.60\% & 12.5 k & 10 & House number images        \\
ImageNet-1    & 37.40\% & 5 k & 10 & Visual recognition images (Tiny ImageNet)         \\
ImageNet-2    & 40.60\% & 5 k & 10 & Visual recognition images (Tiny ImageNet)       \\
ImageNet-3    & 34.20\% & 5 k & 10 & Visual recognition images (Tiny ImageNet)       \\
ImageNet-4    & 36.20\% & 5 k & 10 & Visual recognition images (Tiny ImageNet)       \\ \bottomrule

\end{tabular}
\label{datasets}
\end{specialtable}


\subsection{Representation Performance of the Class Separability Measure $M_{Sep}$}
Here, we show in experiments that how well our first quality measure $M_{sep}$ represents class separability, compared to simple but popular classifiers and the existing data quality~measures.

\subsubsection{Correlation with Classifier Accuracy}
To demonstrate how well $M_{sep}$ represents the class separability of given datasets, we compared the absolute value of Pearson correlation and Spearman rank correlation between quality measures $M_{sep}$, $N1$, $N3$, $F1$, and $CSG$ to the prediction accuracy of three classification models: perceptron, MLP-1, and MLP-2. Table~\ref{table:measure_comparison} summarizes the results. 

In the case of the perceptron, $M_{sep}$ has a similar Pearson correlation with the shortest computation time to the $N1$ and $N3$ which have the highest correlation with the accuracy of classifiers. Furthermore, $M_{sep}$ and $F1$ have the highest Spearman rank correlation. This is because $M_{sep}$ and $F1$ measure linear separability that is essentially the information captured by the linear classifier, the perceptron in our case. In the case of MLP-1 and MLP-2, $M_{sep}$ also showed a sufficiently high correlation with classification accuracy although it is slightly lower in Pearson correlation compared to the case of the perceptron. On the other hand, $CSG$ does not seem to have noticeable benefits considering its computation time. This is because $CSG$ is affected by an embedding network which requires a large amount of training time.

In summary, the result shows that our measure $M_{sep}$ can capture separability of data as good as the existing data quality measures, while reducing computation time significantly.

\begin{specialtable}[H]

\caption{The absolute Pearson and Spearman rank correlation between the quality measures and the accuracy of three classifiers on the ten image datasets (MNIST
, CIFAR10
, notMNIST, Linnaeus, STL10
, SVHN
, ImageNet-1, ImageNet-2, ImageNet-3, and ImageNet-4). The computation time of our method $M_{sep}$ is the fastest.}
\setlength{\tabcolsep}{3.5mm}
\begin{tabular}{ccccc}
\toprule
\textbf{Classifier}                      & \textbf{Quality Measure}   & \textbf{Pearson Corr.}  & \textbf{Spearman Corr.} & \textbf{Time (s)}  \\ \midrule
\multirow{5}{*}{Perceptron}     & $F1$          & 0.9386         & 0.7697           & 1253         \\
                                & $N1$          & 0.9889         & 0.7333            & 5104        \\
                                & $N3$          & 0.9858         & 0.7333            & 9858         \\
                                & $CSG$          & 0.9452         & 0.8182            &  23,711         \\
                                & $M_{sep}$ (ours)& 0.9693         & 0.8061            & 354         \\  \midrule

\multirow{5}{*}{MLP-1}          & $F1$              & 0.9039         & 0.3455            & 1253         \\
                                & $N1$              & 0.9959      & 0.9758            & 5104         \\
                                & $N3$              & 0.9961         & 0.9030            & 9858         \\
                                & $CSG$             & 0.9295         & 0.5879            &  23,711         \\
                                & $M_{sep}$ (ours)  & 0.9261         & 0.3818            & 354         \\  \midrule

\multirow{5}{*}{MLP-2}          & $F1$          & 0.8855              & 0.3455            & 1253         \\
                                & $N1$          & 0.9908        & 0.9273            & 5104        \\
                                & $N3$          & 0.9912        & 0.8788            & 9858         \\
                                & $CSG$          & 0.9127       & 0.5879            &  23,711         \\
                                & $M_{sep}$ (ours) & 0.9117     & 0.4303            & 354        \\ \bottomrule
\end{tabular}
\label{table:measure_comparison}
\end{specialtable}

\subsubsection{Correlation with Other Quality Measures}
In order to check if our suggested data quality measure $M_{sep}$ is compatible with the existing ones in quality, and therefore ours can be a faster alternative to the existing data quality measures, we computed the Pearson correlation between $F1$, $N1$, $N3$, $CSG$, and $M_{sep}$. The results are summarized in Table~\ref{table:measure_correlation}. 
Our measure $M_{sep}$ showed a high correlation with all four existing measures $F1$, $N1$, $N3$, and $CSG$, indicating that $M_{sep}$ is able to capture the data quality information represented by $F1$, $N1$, $N3$, and $CSG$.

\begin{specialtable}[H]
\caption{The absolute Pearson correlation between $M_{sep}$ and other quality measures.}
\setlength{\tabcolsep}{5.3mm}
\begin{tabular}{cccccc}
\toprule
 & \boldmath{$M_{sep}$ (Ours)}   & \boldmath{$F1$}      & \boldmath{$N1$}     & \boldmath{$N3$} & \boldmath{$CSG$}      \\ 
\midrule
$M_{sep}$ (ours) & 1.0000 & 0.9673   & 0.9322  & 0.9245 & 0.9199\\ 
$F1$   & 0.9673  & 1.000   & 0.8909 & 0.8879 & 0.8806\\ 
$N1$   & 0.9322 & 0.8909 & 1.0000   & 0.9988  & 0.9400\\ 
$N3$   & 0.9245 & 0.8879 & 0.9988  & 1.000  & 0.9417\\ 
$CSG$   & 0.9199 & 0.8806 & 0.9400  & 0.9417  & 1.0000\\ 
\bottomrule
\end{tabular}
\label{table:measure_correlation}
\end{specialtable}

\subsubsection{Computation Time}
As mentioned above, our quality measure $M_{sep}$ represents the class separability well, but much faster in computation than $F1$, $N1$, $N3$, and $CSG$. Here, we show how the computation time changes according to data dimension and sample sizes, in order to show that our suggested data quality measure can be used for many big-data situations. 

\begin{figure}[H]
\includegraphics[width=0.9\linewidth]{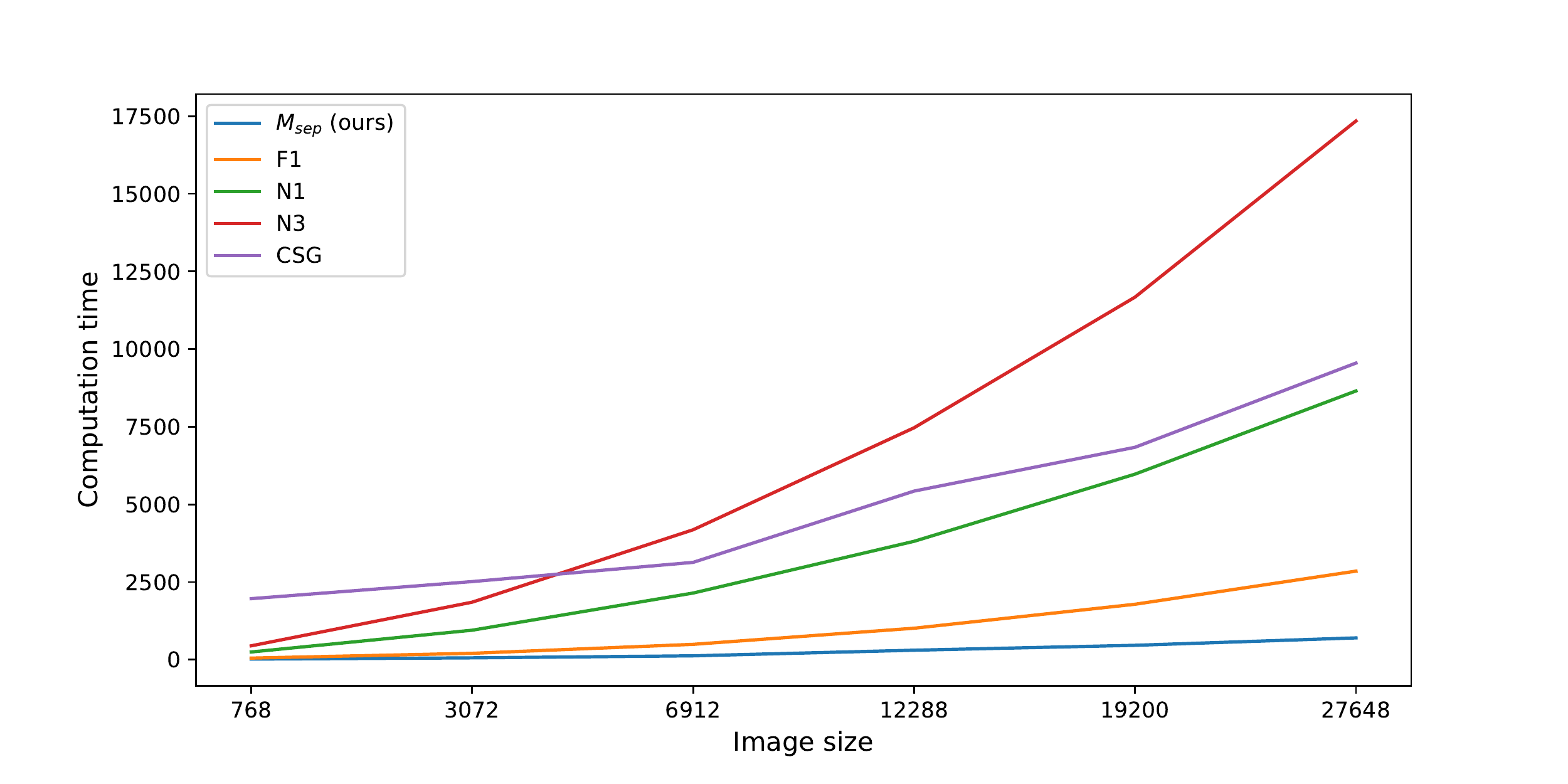}
\caption{Data dimension vs. computation time (CIFAR10).}
\label{fig:time_dim_plot}
\end{figure}

The computation time according to the data dimension is shown in Figure~\ref{fig:time_dim_plot} and \mbox{Table~\ref{fig:time_dim_table}}.
In all dimensions, our measure $M_{sep}$ was on average 3.8 times faster than $F1$, 13.1~times faster than $N1$, 25.9 times faster than $N3$, and 17.7 times faster than $CSG$. Since the $N1$, $N3$, and $CSG$ have to calculate the MST and to train a 1NN classifier and embedding networks, respectively, it is inevitable that they would take a large amount of computation time (see more details in Sections~\ref{subsec:descriptor} and \ref{subsec:graph}). On the other hand, since $M_{sep}$ utilizes random projection and bootstrapping to avoid eigenvalue decomposition problem and to deal with the big-data situations, the computation time of $M_{sep}$ is shortest in all cases.

\vspace{-6pt}

\begin{specialtable}[H]
\caption{Data dimension vs. computation time (CIFAR10) in detail (the values in the table represent~seconds).}
\setlength{\tabcolsep}{2.1mm}
\begin{tabular}{cccccccc}
\toprule
\textbf{Image Size} &    \textbf{Dimension}   & \boldmath{$F1$}     &  \boldmath{$N1$           } &  \boldmath{$N3$ }       &  \boldmath{$CSG$}     &  \boldmath{$M_{sep}$} \textbf{(Ours)}   & \textbf{Speedup}      \\
\midrule
$16 \times 16 \times 3 $ & 768 & 49     & 246         & 445   & 1963 & 17  
& $\times 2.9 \sim 115.5$\\
$32 \times 32 \times 3 $ & 3072 & 203     & 947          & 1850 & 2513  & 57    &  $\times 3.6 \sim 44.1$\\
$48 \times 48 \times 3 $ & 6912 & 492    & 2144         & 4182  &3132 & 122  & $\times 4.0 \sim 34.3$\\
$64 \times 64 \times 3 $& 12,288 & 1011  & 3810          & 7466   & 5427  & 303  & $\times 3.3 \sim 24.6$\\
$80 \times 80 \times 3 $ & 19,200 & 1783   & 5973          & 11,674   & 6838  & 461 & $\times 3.9 \sim 25.3$\\
$96 \times 96 \times 3 $& 27,648 & 2850   & 8652  & 17,346     & 9550   & 700 & $\times 4.1 \sim 24.8$\\ \bottomrule
\end{tabular}

\label{fig:time_dim_table}
\end{specialtable}

\begin{figure}[H]
\includegraphics[width=0.9\linewidth]{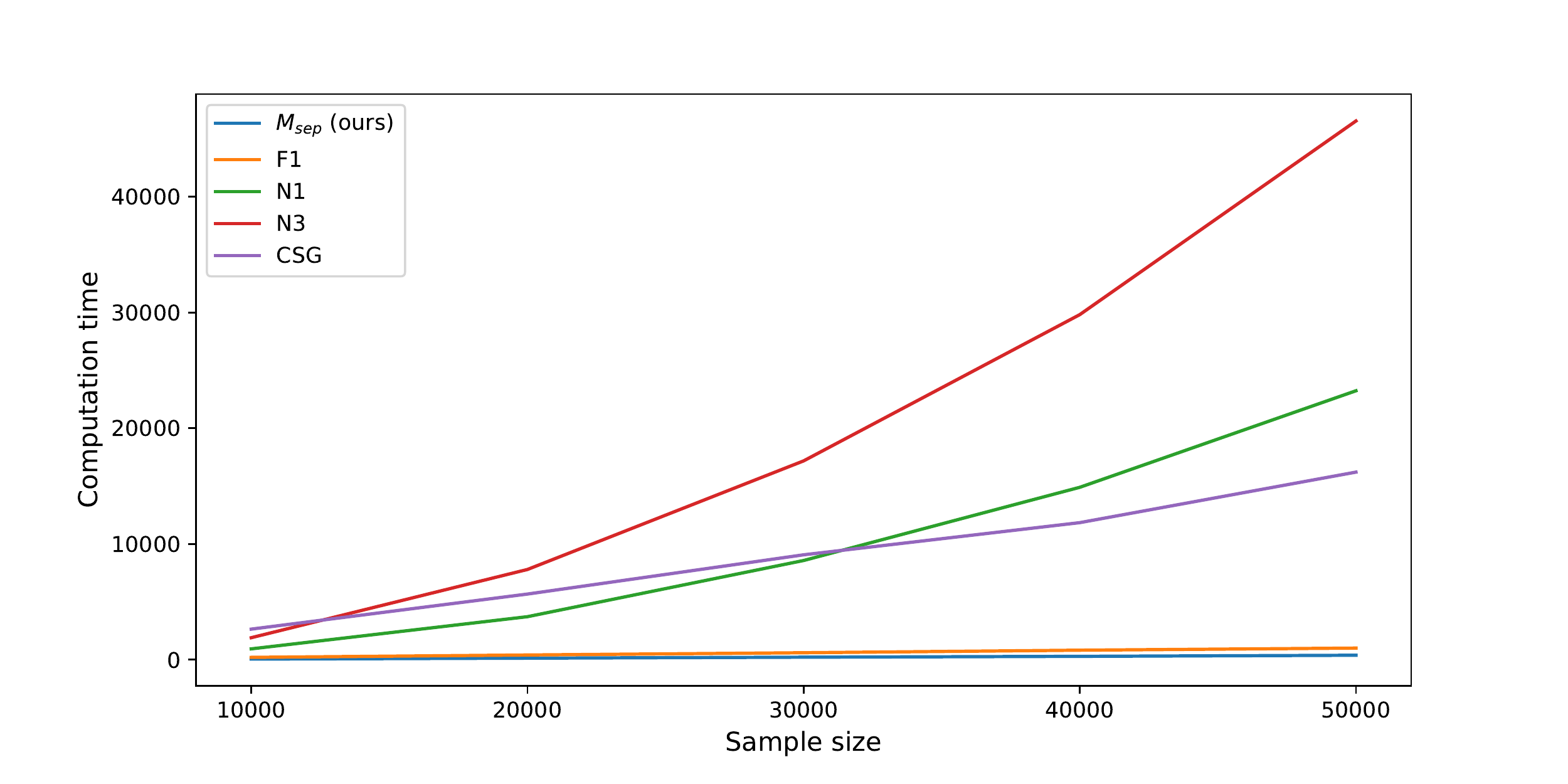}
\caption{Sample size vs. computation time (CIFAR10).}
\label{fig:sample_size_plot}
\end{figure}

Figure~\ref{fig:sample_size_plot} and Table~\ref{table:sample_size_table} show how computation time changes for various sample sizes. Our measure $M_{sep}$ was on average 2.8 times faster than $F1$, 47.0 times faster than $N1$, 94.5 times faster than $N3$, and 41.6 times faster than $CSG$. $N1$ and $N3$ show extremely increasing computation time with respect to the sample size, which is not suitable for large-scale high-dimensional datasets.

All the above results show that our measure $M_{sep}$ is suitable for the big-data situations and compatible with other well-accepted data quality measures.

\vspace{-6pt}
\begin{specialtable}[H]
\caption{Sample size vs. computation time (CIFAR10) in detail (the values in the table represent seconds).}
\setlength{\tabcolsep}{3mm}
\begin{tabular}{ccccccc}
\toprule
\textbf{Sample Size}                & \boldmath{$F1$}     & \boldmath{$N1$ }           & \boldmath{$N3$}     & \boldmath{$CSG$}       & \boldmath{$M_{sep}$} \textbf{(Ours)} & \textbf{Speedup}
\\ \midrule
10,000   & 205  & 939   & 1904  & 2645 & 56 & $\times 3.7 \sim 47.2$ \\
20,000  & 411  & 3719  & 7797 & 5674  & 136   & $\times 3.0 \sim 57.3$\\
30,000  & 605   & 8572 & 17,177   & 9065  & 220 & $\times 2.8 \sim 78.1$\\
40,000 & 825  & 14,896  & 29,813  & 11,843  & 293    & $\times 2.8 \sim 101.8$\\
50,000 & 1006  & 23,235  & 46,541  & 16,208 & 387 & $\times 2.6 \sim 120.3$ \\ \bottomrule
\end{tabular}

\label{table:sample_size_table}
\end{specialtable}

\subsubsection{Comparison to Exact Computation}
\skl{In Section~\ref{sec:proposed_measures}, we proposed to use random projections and bootstrapping for fast approximation of the solution of \eqref{eq:eigen_value}, which can be computed exactly as an eigenvalue. 
Here, we compare the values of $M_{sep}$ using the proposed approximate computation (denoted by ``Approx'') and the exact computation (denoted by ``Exact'') due to an eigensolver in the Python scipy package.
One thing is that, since we use only Gaussian random vectors for projection, it is likely that they may not match the true eigenvectors; therefore, the approximated quantity would differ from the exact value. However, we found that the approximate quantities match well the exact values in their correlation, as indicated in Table~\ref{table:eig_vs_rp}, and, therefore, can be used for fast comparison of data quality of high-dimensional large-scale datasets.}


\begin{specialtable}[H]
\caption{Comparison of Exact and Approx values and their correlations. Pearson and Spearman are the Pearson and Spearman rank correlation between the Exact and Approx.} 

\setlength{\tabcolsep}{16.5mm}
\begin{tabular}{ccc}
\toprule
 & \boldmath{$Exact$} & \boldmath{$Approx$} \\ \midrule
MNIST & 0.1550 & 0.0535\\
CIFAR10 & 0.0331 & 0.0173 \\
notMNIST & 0.2123 & 0.0625 \\
Linnaeus & 0.0250 & 0.0118 \\
STL10 & 0.0695 & 0.0207 \\
SVHN & 0.0004 & 0.0010 \\
ImageNet-1 & 0.0062 & 0.0131 \\
ImageNet-2 & 0.0293 & 0.0145 \\
ImageNet-3 & 0.0191 & 0.0125 \\
ImageNet-4 & 0.0092 & 0.0076 \\ \midrule
Pearson & \multicolumn{2}{c}{0.9847} \\
Spearman & \multicolumn{2}{c}{0.9152} \\ \bottomrule
\end{tabular}
\label{table:eig_vs_rp}
\end{specialtable}


\subsection{Class-Wise In-Class Variability Measure, $M_{Var_i}$}

In fact, many of the existing data quality measures are designed to measure the difficulty of classification for a given dataset. However, we believe that the in-class variability of data must be considered as another important factor of data quality. One example to show the importance and usefulness of our in-class variability measure $M_{var_i}$ is the generalization performance of a classifier. 

The generalization performance of the learning model is an important consideration especially in mission-critical AI-augmented systems. There are many possible reasons causing low generalization, and overfitting is one of the troublemakers. Although we have techniques to alleviate overfitting, e.g., checking the learning curve, regularization~\cite{Srivastava_2014, Ioffe_2015}, and ensemble learning~\cite{Opitz_1999}, it is critical to check if there is an issue in data to begin with which may lead to any inductive bias.
For example, a very small value of $M_{var_i}$ in a class compared to the others would indicate a lack of variability in the class, which can lead to low generalization due to, e.g., unchecked input noise, background signal, object occlusion, and angle/brightness/contrast differences during training. On the other hand, the overly large $M_{var_i}$ may indicate outliers or even mislabeled data points likely to incur unwanted inductive bias in training.

\begin{figure}[H]
\includegraphics[width=1.0\linewidth]{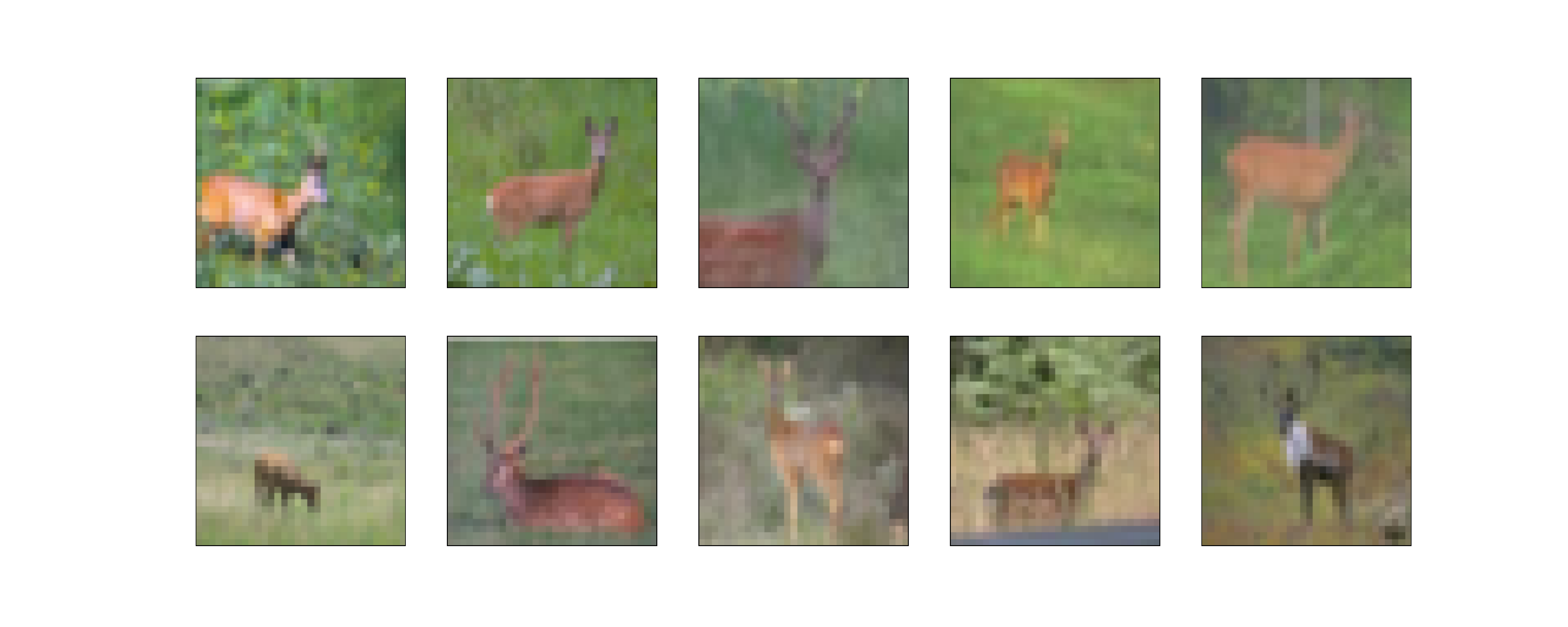}
\caption{Ten similar selected images in the deer class on degraded-CIFAR10. Images with high similarity were selected using cosine similarity.} 

\label{fig:similar_within_class}
\end{figure}

To show the importance and usefulness of in-class variability, we created a degraded version of CIFAR10 (denoted by degraded-CIFAR10) by reducing the variability of a specific class. The degraded-CIFAR10 is created by the following procedure. First, we chose an image of the deer class in the training data, then selected the nine mostly similar images in the angular distance to the chosen one. Figure~\ref{fig:similar_within_class} shows the total ten images selected by the above procedure that have similar backgrounds and shapes. 
Next, we created 1000~images by sampling with replacement from the ten images, while adding random Gaussian noise with zero mean and unit variance, and we replaced the original deer class data with sampled degraded deer class data.

Table~\ref{table:within-class variance} shows that the value of $M_{var_i}$ is significantly small on the degraded deer class compared to the other classes. That is, it can capture small in-class variability. In contrast, Table~\ref{table:toy_measure} shows that the existing quality measures $F1$, $N1$, $N3$, and $CSG$ may not be enough to signify the degradation of the dataset. As we can see, all quality measures indicate that class separability increased in degraded-CIFAR10 compared to the original CIFAR10; however, the test accuracy from MLP-2 decreased. This is because the reduction in in-class variability is very likely to decrease the generalization performance. Therefore, class separability measures can deliver incorrect information regarding data quality in terms of in-class variability, which can be a critical problem for generating a trustworthy dataset or training a trustworthy model.



\begin{specialtable}[H]
\caption{Our in-class variability measure for the degraded-CIFAR10 dataset. A class with a smaller value than other classes has a lower variability.} 
\setlength{\tabcolsep}{15mm}
\begin{tabular}{cc}\toprule
\backslashbox[25em]{\textbf{Class}}{\textbf{Measure}} & \boldmath{$M_{var_i}$ $\times$ 1000} \\ \midrule
Airplane & 0.1557 \\
Automobile & 0.2069 \\
Bird & 0.1394 \\
Cat & 0.1803 \\
Deer & \textbf{0.0123} \\  
Dog & 0.1830 \\
Frog & 0.1344 \\
Horse & 0.1775 \\
Ship & 0.1472 \\
Truck & 0.1997 \\ \bottomrule

\end{tabular}
\label{table:within-class variance}
\end{specialtable}

\begin{specialtable}[H]
\caption{Quality measures on the degraded-CIFAR10 dataset. The existing quality measures $F1$, $N1$, $N3$, and $CSG$ only capture the class separability and fail to capture the degradation. Lower values of $N1$, $N3$, and $CSG$ represent higher class separability, whereas lower values of $F1$ represent lower class separability. The test accuracy is from MLP-2 trained with original and degraded-CIFAR10, respectively, and tested on the original CIFAR10 test data.}
\setlength{\tabcolsep}{2mm}
\begin{tabular}{cccccc}\toprule
\backslashbox[14em]{\textbf{Data}}{\textbf{Quality Measure}}&
\boldmath{{$F1$}}&\boldmath{{$N1$}} &\boldmath{{$N3$}}&\boldmath{{$CSG$}}&\boldmath{{$Test\ Accuracy\ (\%)$}}\\\midrule
Original CIFAR10 & 0.2213 & 0.7909 & 0.7065& 0.7030 & 42.84 \\
Degraded-CIFAR10 & 0.2698 & 0.7035 & 0.6096& 0.6049 & 41.28\\ \bottomrule
\end{tabular}
\label{table:toy_measure}
\end{specialtable}


As we showed above, the small value of $M_{var_i}$ of a specific class represents that similar images do exist in the invested class, which can lead to low generalization performance of classifiers. Suppose we have generated a dataset for an autonomous driving object classification task. The dataset has been revealed that it has a high class separability through various quality measures. Moreover, the training accuracy was also high. Therefore, one may expect high generalization performance. Unfortunately, the exact opposite can happen. If the variability in the specific class is small as in the degraded-CIFAR10 example above, high generalization performance cannot be expected. For instance, if a car with new colors and new shapes that have never been trained is given as an input to the model, the probability of properly classifying the car will be low. This example indicates that in-class variability plays an important role in data quality evaluation.

\subsection{Quality Ranking Using $M_{Sep}$ and $M_{Var}$}
\begin{figure}[H]
\includegraphics[width=0.9\linewidth]{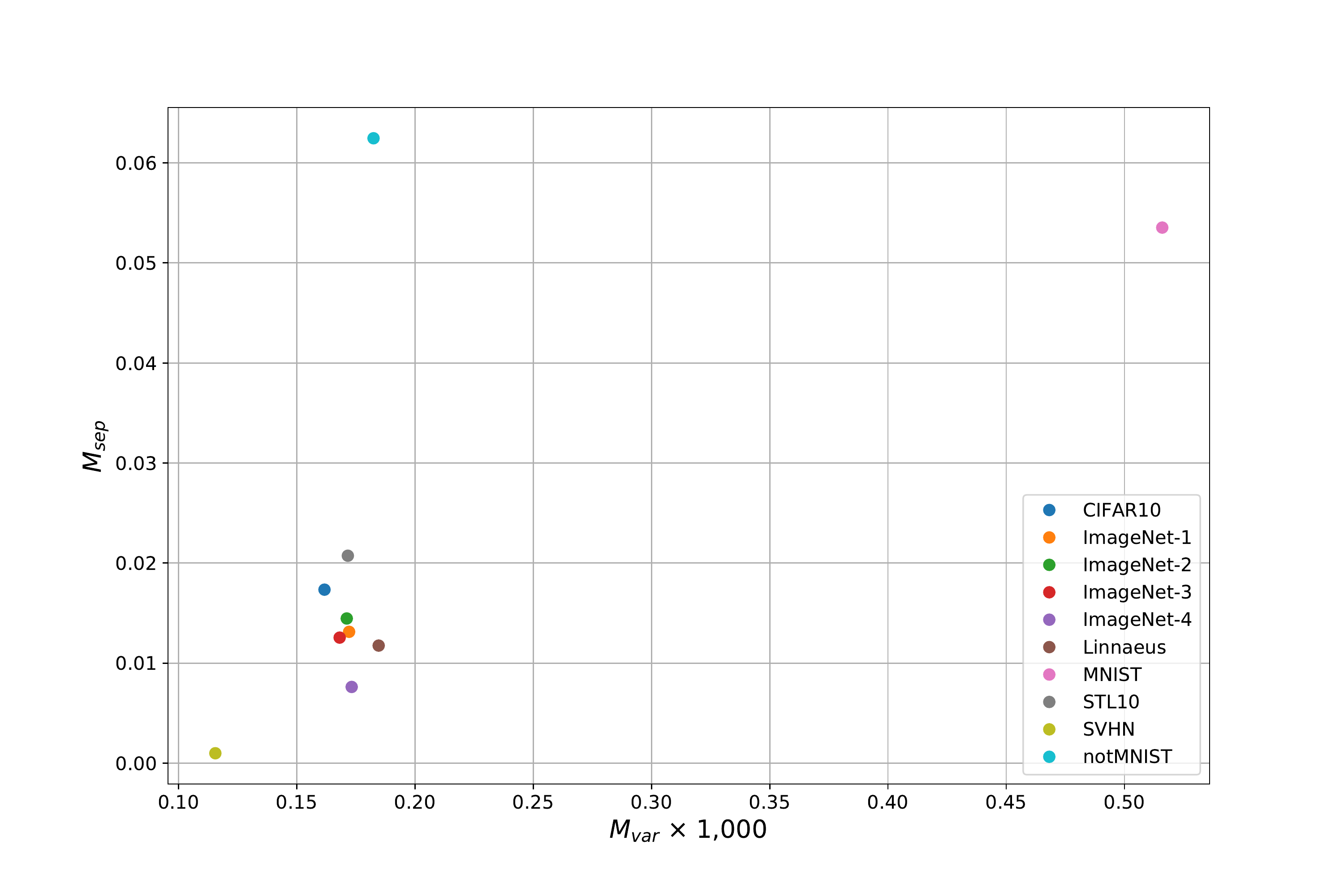}
\caption{{Data quality plot} using the two proposed quality measures.}
\label{fig:scatter_plot}
\end{figure}

\skl{As we mentioned before, quality measures $M_{sep}$ and $M_{var}$ can be compared among different datasets. The class separability $M_{sep}$ represents the relative difficulty of linear classification, and the overall in-class variability $M_{var}$ represents the average variability of data points in classes.

Figure~\ref{fig:scatter_plot} shows a data quality comparison plot of datasets in our experiments.}
\skl{ The direction towards the lower-left corner indicates lower class separability and lower in-class variability, and the upper-right direction is for higher class separability and higher in-class variability. 
According to the plot, the MNIST and the notMNIST dataset show very high linear separability compared to other datasets, indicating that their classification might be easier than the other datasets.
The SVHN dataset is at the lower-left corner, indicating low linear separability and possible redundancy issues (this could be just the reflection of the fact that many SVHN images contain changing digits but the same backgrounds). The four ImageNet datasets, Linnaeus, CIFAR10 and STL10 have similar class separability and in-class variability values. This appears to be understandable considering their similar data construction designed for object recognition.}

\section{Conclusions}
In this paper, we proposed data quality measures $M_{sep}$, $M_{var}$ and $M_{var_i}$, which can be applied efficiently on large-scale high-dimensional datasets. Our measures are estimated using random projection and bootstrapping and therefore can be applied efficiently on large-scale high-dimensional data. We showed that $M_{sep}$ can be used as a good alternative to the existing data quality measures capturing class separability, while reducing their computational overhead significantly. In addition, $M_{var}$ and $M_{var_i}$ measures in-class variability, which is another important factor to avoid unwanted inductive bias in trained models. 
\vspace{6pt}



\authorcontributions{Conceptualization, H.C. and S.L.; methodology, H.C. and S.L.; validation, H.C.; writing-original draft preparation, H.C.; writing-review and editing, S.L.; supervision, S.L. Both authors have read and agreed to the published version of the manuscript.}

\funding{This research was supported by Basic Science Research Program through the National Research Foundation of Korea(NRF) funded by the Ministry of Education(2018R1D1A1B07051383), and also by the MSIT(Ministry of Science and ICT), Korea, under the ITRC(Information Technology Research Center) support program(IITP-2020-0-01749) supervised by the IITP(Institute of Information
\& Communications Technology Planning \& Evaluation).}





\dataavailability{The data presented in this study are openly available: MNIST (\url{http://yann.lecun.com/exdb/mnist/}), notMNIST (\url{https://www.kaggle.com/lubaroli/notmnist}), Linnaeus (\url{http://chaladze.com/l5/}), CIFAR10 (\url{https://www.cs.toronto.edu/~kriz/cifar.html}), STL10 (\url{https://ai.stanford.edu/~acoates/stl10/}), SVHN (\url{http://ufldl.stanford.edu/housenumbers/}), and ImageNet (\url{https://www.kaggle.com/c/tiny-imagenet}).}  

\conflictsofinterest{The authors declare no conflict of interest.} 

\end{paracol}
\reftitle{References}





\typeout{get arXiv to do 4 passes: Label(s) may have changed. Rerun}
\end{document}